\documentclass{easychair}

\usepackage{amsmath}
\usepackage{enumitem}
\usepackage{tikz}
\usepackage[]{algorithm2e}
\usepackage{hyphenat}
\usetikzlibrary{matrix}

\usepackage{doc}

\newcommand{\Comment}[1]{}

\title{Designing Game of Theorems}

\author{
Yutaka Nagashima
}

\institute{
  CIIRC, CTU, Prague / The University of Innsbruck, Innsbruck\\
  \email{first\_name.last\_name@cvut.cz}
 }

\authorrunning{Yutaka Nagashima}

\titlerunning{Towards Smart Proof Search for Isabelle}
\begin{document}

\maketitle

\begin{abstract}
``Theorem proving is similar to the game of Go. So, we can probably improve our provers using deep learning, like DeepMind built the super-human computer Go program, AlphaGo \cite{alphaGo}.'' Such optimism has been observed among participants of AITP2017. But is theorem proving really similar to Go? In this paper, we first identify the similarities and differences between them and then propose a system in which various provers keep competing against each other and changing themselves until they prove conjectures provided by users.

\end{abstract}



%
%

\pagestyle{empty}

\section{The Game of Go and Theorem Proving}

\paragraph{Formally defined rules [similarity].}
Both the games of Go and theorem proving have algorithms to evaluate the results. In the game of Go, one can judge the result of each game when it is over by counting the stones and spaces for each player, and no ambiguity is left in deciding the result of a game. In theorem proving, when one finds a proof, others can systematically check if the alleged proof is a valid proof or not.

\paragraph{Expressive power of the system [difference].}
Even though both systems are based on a set of simple rules, the expressive power of these systems differ. Depending on the underlying logics, a theorem proving task can involve advanced concepts such as abstraction, universal quantification, existential quantification, and polymorphism, which Go scores cannot express natively. This is especially true for more expressive logics such as classical higher-order logics or variants of calculus of constructions, where stronger proof automation is needed.

\paragraph{Amount of available training data [difference].}
Some theorem proving researchers boast that they have large proof corpora. For example, the Isabelle community has the Archive of Formal Proofs (AFPs) \cite{AFP}, consisting of more than 1.5 millions of lines of code and 100 thousands lemmas. Unfortunately, even though these proof corpora are large for the small community of theorem provers, they are small compared to the data deployed in other domains.

\paragraph{Preference towards small data [difference].}
The size of the community is not the only reason of small data available in the theorem proving community: logicians and mathematicians have developed expressive logics to describe general ideas in a concise manner. Combined with the trade-off between proof automation and expressive power of underlying logic, this is doubly unfortunate: the more expressive logic we use, the less proof automation we have, but the more expressive the logic is, the less training data we can expect, which makes it hard to improve the proof automation for expressive logics using machine learning techniques.

\paragraph{Self-playability [similarity/difference].}
One might suspect that large data are not necessary to develop a powerful proof automation tool using machine learning. After all, DeepMind has made AlphaGo Zero \cite{alphaGoZero} stronger than any previous versions of AlphaGo via self-play without using data from human games. Unfortunately, even though both Go and theorem proving are based on clearly defined rules, 
theorem proving is not a two-player game by default. In the rest of this paper, we propose an approach to introducing self-playability to theorem proving.

\section{The Design of Self-playable Games of Theorem Proving}
One straightforward design of self-playable games of theorem proving is as follows: 
(1) prepare a set of proof obligations from existing proof corpora,
(2) let two competing provers try to prove these proof obligations,
(3) count how many obligations each prover discharges,
(4) consider the prover that solves more obligations as the winner, and the other one as the loser.
We can use this naive approach as a part of reinforcement learning or evolutionary computation to optimize our provers for proof obligations that have already been proved. However, this approach is probably not powerful enough to improve provers for conjectures that are significantly different from the theorems in the training data 

For example, let us assume that we enhance our prover, say \verb|P|, via self-play using 100 theorems and their proofs in the AFPs. Since we already know how to prove these theorems, we can improve \verb|P|, so that \verb|P| can prove all of the 100 theorems within a reasonable time-out. However, when we try to improve \verb|P| to discharge a new conjecture, say Goldbach's conjecture, we will find ourselves at a loss of training data: Currently, there is no known proofs of Goldbach's conjecture or auxiliary lemmas that are verified to be useful to prove Goldbach's conjecture. 

If we add Goldbach's conjecture to the above dataset, the improvement via self-play would saturate after producing a prover that can discharge the 100 theorems from the AFP but not Goldbach's conjecture:
since the gap between the theorems from the AFPs and Goldbach's conjecture is too large, minor mutations to \verb|P|'s variants cannot produce a useful observable difference in the result of the game. What we need here is a mechanism to produce conjectures that we can reasonably expect to be useful to prove our target conjecture (Goldbach's conjecture in this example) but not too difficult for our current prover \verb|P|.

Therefore, we propose to \textit{treat conjecturing and proof search as one problem}. Of course, we cannot be 100\% sure which conjecture is useful to train our prover for Goldbach's conjecture, since nobody has proved it yet. But if we consider a heuristic proof search as the exploration of an and-or tree, we can estimate how important each node in the tree is from the search heuristics of the prover. 
Furthermore, given a long time-out, we can expect that the prover can discharge some of emerging subgoals, even if the prover cannot discharge the root-node, which corresponds to the target conjecture (Goldbach's conjecture, in this example). 

Our idea is to \textit{use these proved subgoals to judge the competence of other versions of prover} \verb|P|. 
First, we produce two versions of our prover \verb|P| by mutation. Let us call them \verb|Pa| and \verb|Pb|, respectively. Using the approach explained above, we let \verb|Pa| produce a dataset \verb|Da| and let \verb|Pb| produce \verb|Db|. Now, we let \verb|Pb| try to prove the theorems in \verb|Da|, and let \verb|Pa| try to prove the theorems in \verb|Db|. 
When \verb|Pa| and \verb|Pb| run out of time, we sum up the estimated values of theorems proved by each prover (\verb|Pa|, for example). Note that it was the opponent prover (\verb|Pb| in this example) that has decided the value of each theorem in each dataset (\verb|Db| in this case) when finding proofs of these subgoals for the first time. The prover that has gained more value is the winner of the game, and the other is the loser. Then, we keep running this game by mutating the winner until we produce a prover that can discharge the target conjecture. Since this process generates new conjectures tagged with their estimated values from the target conjecture in each iteration, we expect this approach continues producing conjectures useful to prove the target conjecture.

We are still in the early stage of the design. We might generalize this idea to n-player games to avoid over-fitting. For the moment, we prefer the design of the game to be irrelevant to any of underlying logics, ML algorithms for search heuristics, and mutation algorithms. 


\section*{Acknowledgement}
This work was supported by the European Regional Development Fund under the project AI\&Reasoning (reg. no. CZ.02.1.01/0.0/0.0/15\_003/0000466).

\label{sect:bib}
\bibliographystyle{plain}
\bibliography{easychair}


\end{document}